\documentclass[10pt,twocolumn,letterpaper]{article}

\usepackage[final]{iccv_template/cvpr}
\usepackage{times}
\usepackage{epsfig}
\usepackage{graphicx}
\usepackage{amsmath}
\usepackage{amssymb}
\usepackage{caption}

\usepackage{booktabs}       
\usepackage{array}
\usepackage{multirow}

\usepackage{siunitx}
\usepackage{xspace}

\usepackage{todonotes}
\usepackage{comment}

\usepackage{amsfonts}       
\usepackage{nicefrac}       
\usepackage{microtype}      
\usepackage[super]{nth}

\usepackage{enumitem}
\usepackage{lipsum}
\usepackage{wrapfig}
\usepackage{tikz}
\usetikzlibrary{calc,spy}

\usepackage{cancel}

\newcommand{\isdraft}{false}

\usepackage[pagebackref=true,breaklinks=true,colorlinks,bookmarks=false]{hyperref}

\usepackage{doi}

\usepackage[capitalize]{cleveref}
\crefname{section}{Sec.}{Secs.}
\Crefname{section}{Section}{Sections}
\Crefname{table}{Table}{Tables}
\crefname{table}{Tab.}{Tabs.}

\def\cvprPaperID{24} 
\def\httilde{\mbox{\tt\raisebox{-.5ex}{\symbol{126}}}}

\begin{document}

\title{`Tax-free' 3DMM Conditional Face Generation}

\author{Yiwen Huang
\and
Zhiqiu Yu
\and
Xinjie Yi
\and
Yue Wang
\and
James Tompkin
\and
Brown University
}
\maketitle


\begin{abstract}
3DMM  conditioned face generation has gained traction due to its well-defined controllability; however, the trade-off is lower sample quality: Previous works such as DiscoFaceGAN and 3D-FM GAN show a significant FID gap compared to the unconditional StyleGAN, suggesting that there is a quality tax to pay for controllability. In this paper, we challenge the assumption that quality and controllability cannot coexist. To pinpoint the previous issues, we mathematically formalize the problem of 3DMM conditioned face generation. Then, we devise simple solutions to the problem under our proposed framework. This results in a new model that effectively removes the quality tax between 3DMM conditioned face GANs and the unconditional StyleGAN.
\end{abstract}
\vspace{-0.25cm}
\section{Introduction}
\label{sec:intro}
\vspace{-0.1cm}

Face image generation has wide application in computer vision and graphics. Among different works in this area, deep learning generative model approaches are especially good at generating high-quality photo-realistic face images~\cite{sg2,sg3}. However, generative models provide limited explicit control over their output due to their unsupervised nature, relying instead on latent space manipulation~\cite{fadernetworks}. On the other hand, parametric models such as 3D Morphable Models (3DMMs) embed facial attributes in a disentangled parameter space, but their results lack photorealism~\cite{paysan20093d}. 

In light of this, researchers have tried to build models that can synthesize high-resolution novel face images with control by combining 3DMM with generative modeling~\cite{dfg,3fg,stylerig,styleflow,gif}. Existing attempts can be roughly divided into two categories: rigging and conditional generation. Rig-based methods attempt to align the 3DMM parameter space with the latent space of a pre-trained generative model~\cite{stylerig,styleflow}. Sample quality is not compromised by controllability; however, controllability is limited by the completeness and disentanglement of the underlying latent space~\cite{stylespace}. Conditional generation methods use the 3DMM when training the generative model~\cite{dfg,3fg,gif}. These offer improved controllability but reduced sample quality since additional constraints are imposed upon the generated samples for 3DMM consistency and disentanglement.

We investigate the family of 3DMM conditional GAN models. Deng~\etal.~state that the quality drop in conditional models is an inevitable tax that we pay for controllability~\cite{dfg}. What causes this tax? We hypothesize that it is caused by overconstraint: that, to achieve consistency with the 3DMM conditioning \emph{and} disentanglement among latent variables, current methods have unnecessary side effects that compromise quality. We challenge the claim of a `quality tax' and show that it can be largely eliminated if the overconstraints can be identified and resolved. To this end, we formalize 3DMM conditioned face generation and identify minimal solutions that satisfy controllability and disentanglement.

\begin{figure}[t]
   \vspace{-0.55cm}
   \centering
   \includegraphics[width=\linewidth]{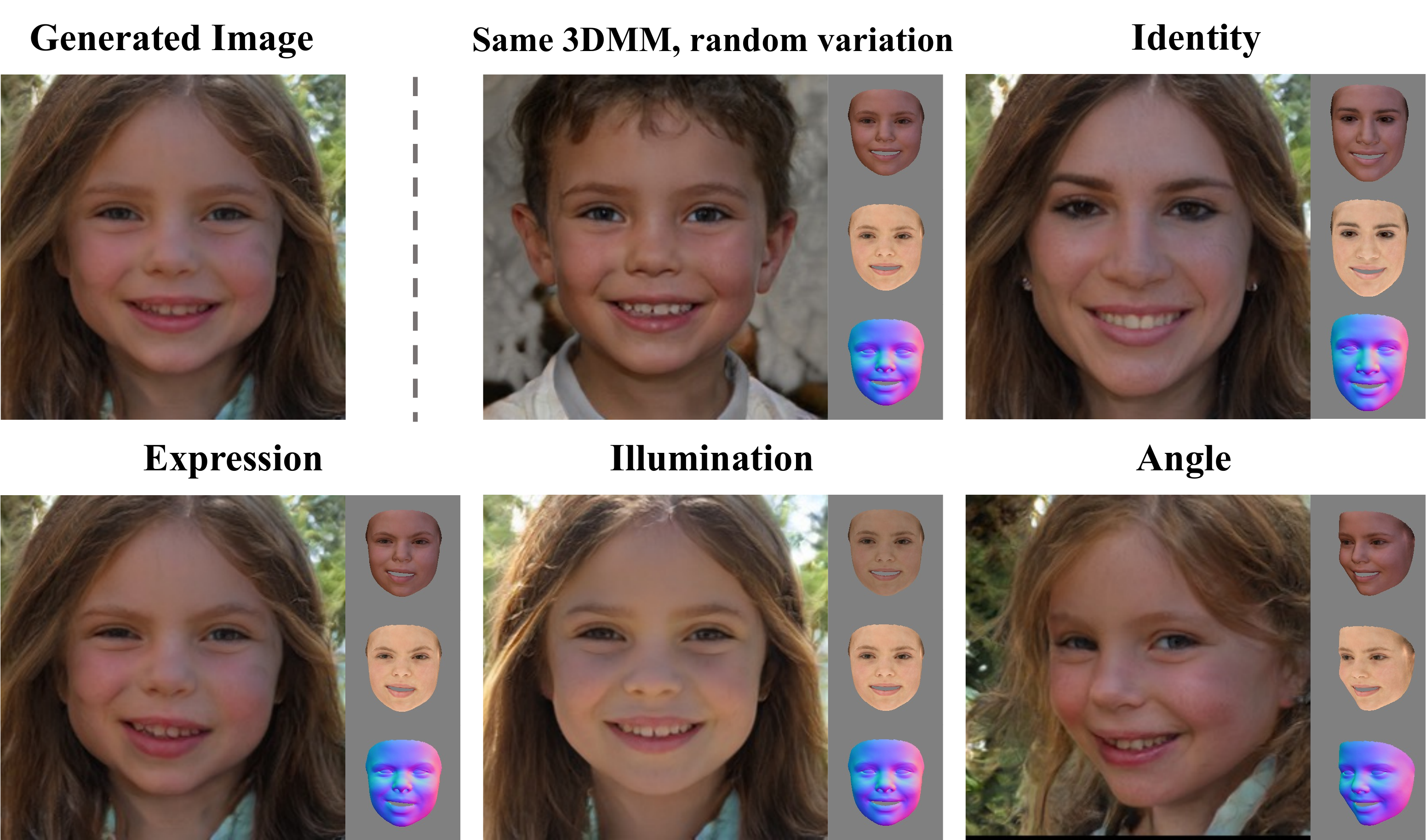}
   \vspace{-0.6cm}
   \caption{3DMM conditioned GANs show reduced image generation quality as a `tax' for their added control. This tax is not inevitable. Our approach produces images of almost equivalent quality to unconditional generation while being at least as disentangled for control.}
   \label{fig:teaser}\vspace{-0.5cm}
\end{figure}

To summarize, our contributions are threefold:
\begin{itemize}[topsep=1pt,parsep=2pt,itemsep=2pt,left=2pt]
\item We propose a mathematical framework for 3DMM conditioned face generation, and unify existing methods under such formulation. This allows us to analyze the consistency and the disentanglement behavior rigorously. 
\item We derive new methods to achieve consistency and disentanglement from our mathematical framework. We show that our methods are both theoretically justified and perform favorably against previous work in practice. 
\item We demonstrate a StyleGAN2-based model trained by our methods that achieves state-of-the-art FID while maintaining the full controllability of 3DMM.
\end{itemize}

\section{Method}
\subsection{Background and Problem Formulation}

We define face images in a dataset $\hat{x} \in \mathcal{X}$.
We also define a 3DMM code vector by $p=\{z_\text{id},z_\text{exp},z_\text{illum},z_\text{angle},z_\text{trans}\}$, a noise vector $z$, and a generator model $G(p,z):\mathcal{P}\times\mathcal{Z}\rightarrow \mathcal{X}$. The goal of conditional generation is to create photorealistic face images $x$ according to $p$ and $z$. For our goal, we can form two related yet distinct objectives: \emph{consistency} and \emph{disentanglement}.

\vspace{-2 ex}
\paragraph{Consistency.}
This objective requires that $x$ is semantically consistent with $p$,~\ie, $p$ dictates the corresponding semantic factors in $x$. We follow the formulation in InfoGAN~\cite{infogan} and formalize the consistency objective as maximizing the mutual information $I(p;x)$ between $p$ and $x$:
\begin{align}
\label{eq:info}
I(p;x)&=H(p)-H(p\mid x) \nonumber \\
&=\mathbb{E}_{x\sim G(p,z)}\left[ \mathbb{E}_{p^\prime\sim P(p\mid x)}\left[\log P(p^\prime\mid x)\right] \right]+H(p) \nonumber \\
&=\mathbb{E}_{p\sim P(p),x\sim G(p,z)}\left[\log P(p\mid x)\right]+H(p)
\end{align}

For 3DMM conditioned face generation, the posterior $P(p|x)$ becomes tractable when the generative distribution $P_g$ becomes sufficiently close to the distribution of real face images. In such case, the posterior is exactly represented by a pretrained face reconstruction model~\cite{recon} that can accurately predict $p$ given $x$, allowing $I(p;x)$ to be directly optimized.

Past works~\cite{dfg, 3fg} propose proxy objectives instead of directly maximizing $I(p;x)$. These objectives maximize $I(p;x)$ up to some deterministic transformation on $p$. We show that directly optimizing the mutual information objective is better than optimizing proxy objectives. Further, as the assumption that $P_g$ is sufficiently close to the real image distribution does not hold in general early in training, we also introduce a progressive blending mechanism.

\vspace{-2ex}
\paragraph{Disentanglement.}
Changing one semantic factor should not interfere with other semantic factors. Let $\mathcal{P}\cup\mathcal{Z}=\{z_0,z_1,\dots,z_n\} $ where $z_i$ denotes the latent code for an independent semantic factor. We formally define disentanglement following Peebles~\etal~\cite{peebles2020hessian}:

\begin{equation}
\label{eq:disentangle}
\frac{\partial^2G}{\partial z_j\partial z_i}=0 \,\,\,\,\,\,\,\,\,\,\,\,\,\, \forall\,i\neq j
\end{equation}

Suppose we define a subset of latent factors that control 3DMM factors; $z_i \in\mathcal{P}$. For these, disentanglement is achieved by construction via the consistency objective. The remaining problem is to disentangle unsupervised factors $z_j\in\mathcal{Z}$ from $z_i \in\mathcal{P}$. Finally, as noted, the disentangling of unsupervised factors $z_j\in\mathcal{Z}$ from each other is an open question~\cite{semisg, disen_nonid} and does not relate to 3DMM conditioning.

In the simplest case where $G$ is a scalar function and each semantic factor $z_i$ is also a scalar, Eq.~\ref{eq:disentangle} indicates that the Hessian matrix $\textbf{H}_G$ is diagonal. In such case, disentanglement can be directly encouraged by a Hessian penalty. However, it is observed that a Hessian penalty has a strong negative impact on image quality (measured by FID~\cite{fid})~\cite{peebles2020hessian} and a solution to this problem is not yet clear.

As we found for consistency, disentanglement is also approximated by proxy objectives in previous work~\cite{dfg, 3fg}. We notice that all such approximations are restrictive; they degrade image quality and rely on hand-designed rules that only work for certain $z_i$. To this end, we propose an alternative approach to the disentanglement problem. We show in the following section that, in practice, disentanglement can be achieved \emph{for free} without any optimization via the inductive bias of a carefully designed network.

\subsection{Consistency via $p$ Rendering \& Estimation}
We maximize Eq.~\ref{eq:info} to enforce semantic consistency between $p$ and $x$. However, there remains a design space of deterministic transformations on $p$ to obtain a more amenable representation for conditioning and optimizing $G$.
To this end, we use a differentiable renderer $\textbf{RDR}$~\cite{recon} to derive a 3DMM representation that aligns with the image space perceptually, and is independent of external factors such as the PCA bases that $p$ depends on. Specifically, we let $\textbf{RDR}$ output the 3DMM rendered image $r$ from $p$, the Lambertian albedo $a$, and the normal map $n$:
\begin{equation}
r,a,n=\textbf{RDR}(p)
\end{equation}

\noindent We define our 3DMM representation `$\text{rep}$' as the Cartesian product of $r$, $a$ and $n$:
%
$\text{rep}(p) = r\times a\times n.$
%
Given the new 3DMM representation, we update Eq.~\ref{eq:info}:
\begin{align}
\label{eq:infonew}
I(\text{rep}(p);x)&=\mathbb{E}_{p\sim P(p),x\sim G(\text{rep}(p),z)}\left[\log P(\text{rep}(p)\mid x)\right]+C
\raisetag{2.5\normalbaselineskip} 
\end{align}

\noindent where $C$ is the constant term $H(\text{rep}(p))$.

\paragraph{Consistency loss.}
Given a pretrained face reconstruction model $\textbf{FR}$~\cite{recon}:$\mathcal{X}\rightarrow\mathcal{P}$, we rewrite Eq.~\ref{eq:infonew} as follows:
\begin{align}
\label{eq:lcons}
\resizebox{\linewidth}{!}{$
\mathcal{L}_\text{consistency}=\mathbb{E}_{p\sim P(p),x\sim G(\text{rep}(p),z)}\left[\left \| \text{rep}(\textbf{FR}(x))-\text{rep}(p) \right \|^{\text{p}}_\text{p}\right].
$}
\raisetag{3\normalbaselineskip} 
\end{align}

\noindent The choice of $\text{p}$ depends on our assumption about the functional form of the posterior. We follow common assumptions and assume Gaussian error, which leads to $\text{p}=2$.~\cite{deeplearning}

\vspace{-2ex}
\paragraph{Progressive blending.}
The posterior $P(p|x)$ can only be represented by $\textbf{FR}$ when $P_g$ is sufficiently close to the real image distribution. 
In early training with Eq.~\ref{eq:lcons}, $x$ is not a realistic image and so $\textbf{FR}(x)$ is nonsensical. To circumvent this problem, we introduce a progressive blending variant of Eq.\ref{eq:lcons}, following the intuition that $r$ is always a close enough approximation of the real face for $\textbf{FR}$:
\begin{align}
\mathcal{L}^*_\text{consistency} &=\mathbb{E}_{p\sim P(p),x\sim G(\text{rep}(p),z)}\left[ d \right] \\
d &= \left \| \text{rep}(\textbf{FR}(\alpha x+(1-\alpha)r(p)))-\text{rep}(p) \right \|^{\text{2}}_\text{2} \nonumber 
\raisetag{3\normalbaselineskip} 
\end{align}

\noindent where $\alpha$ is a scalar that grows linearly from 0 to 1 in the first $k$ training images. We empirically find that this simple strategy is sufficient to solve the intractable posterior problem early in the training.

\subsection{Structurally Disentangled Conditioning}
Next, we discuss how we use $\text{rep}(p)$ to condition $G$. We generate per-layer conditioning feature maps $c=\{c_1,\dots,c_l\}$ using an encoder $E$, and inject each $c_i$ into the corresponding layer of the synthesis network as an auxiliary input. We show that our conditioning method approximates Eq.~\ref{eq:disentangle} without supervision~\cite{dfg,3fg}, achieving disentanglement \emph{for free} as an inductive bias of the network architecture.

\vspace{-2ex}
\paragraph{Feature injection.}
We extend each synthesis layer $l_i$ to take an auxiliary input $c_{n-i}$ where $n$ is the number of layers in the synthesis network. The synthesis layer in~\cite{sg2} is implemented by a stylized convolution where each channel $f_j$ of the input feature maps $f$ is scaled by $s_{ij}$. The per-layer scaling vector $s_i=\{s_{ij}\,\forall j\}$ is computed from the style vector $w_i$ via an affine transformation. We note that the injected feature maps $c_{n-i}$ need to be handled separately for stylization. This is because $c_{n-i}$ is essentially an embedding of $\mathcal{P}$ while $w_i$ is an embedding of $\mathcal{Z}$. It is clear that $\mathcal{P}$ is not controlled by $\mathcal{Z}$ and therefore $c_{n-i}$ should not be subject to $w_i$. To this end, we simply fix the scaling of each channel of $c_{n-i}$ to 1 for stylization.

\paragraph{Disentanglement analysis.}
To simplify analysis, we omit various details from the StyleGAN2~\cite{sg2} generator (weight demodulation, noise injection, equalized learning rate,~\etc). 
We formulate each layer $l_i$ of the synthesis network as:
\begin{equation}
l_i(p,z)=\textbf{W}_i\ast[c_{n-i}(p);s_i(z)\odot\sigma(l_{i-1}(p,z))]+\textbf{B}_i
\end{equation}

\noindent $\textbf{W}_i$ is the weight tensor of $l_i$, $\textbf{B}_i$ is the bias tensor of $l_i$, $\ast$ denotes convolution, $\odot$ denotes the Hadamard product, and $\sigma$ is the activation function. There are two terms in $l_i$ that depend on $p$: $c_{n-i}$ and $\sigma(l_{i-1})$. First, we analyze disentanglement~\wrt $c_{n-i}$:
\begin{align}
\frac{\partial^2l_i}{\partial_{z}\partial c_{n-i}} &=\frac{\partial}{\partial_{z}}\left( \frac{\partial}{\partial c_{n-i}}\left(\textbf{W}_i\ast[c_{n-i};s_i\odot\sigma(l_{i-1})]+\textbf{B}_i\right )  \right) \nonumber \\
&=\frac{\partial}{\partial_{z}}\left(\textbf{W}_i\ast \frac{\partial}{\partial c_{n-i}} [c_{n-i};s_i\odot\sigma(l_{i-1})] \right) \nonumber \\
&=\frac{\partial}{\partial_{z}}\left(\textbf{W}_i\ast[I;0]\right) \nonumber \\
&=0
\raisetag{3\normalbaselineskip} 
\end{align}

\noindent We see that variation in $c_{n-i}$ is perfectly disentangled from variation in $z$, therefore any non-zero $\frac{\partial^2l_i}{\partial_{z}\partial_p}$ must be the result of variation in $\sigma(l_{i-1})$:

\begin{equation}
\begin{aligned}[t]
&\,\,\,\,\,\,\,\,\frac{\partial^2l_i}{\partial z\partial p}=\frac{\partial^2l_i}{\partial z\partial\sigma(l_{i-1})} \frac{\partial\sigma(l_{i-1})}{\partial p} \\
&=\left(\textbf{W}_i\ast \left[0;\frac{\partial s_i}{\partial z}\right] \right)\frac{\partial\sigma(l_{i-1})}{\partial p}
\end{aligned}
\end{equation}

\noindent We examine the behavior of variation in $p$:
\begin{equation}
\resizebox{0.42\textwidth}{!}{$
\begin{aligned}[t]
&\,\,\,\,\,\,\,\,\frac{\partial\sigma(l_{i-1})}{\partial p}=\frac{\partial\sigma(l_{i-1})}{\partial l_{i-1}}\frac{\partial l_{i-1}}{\partial p} \\
&=\frac{\partial\sigma(l_{i-1})}{\partial l_{i-1}}\left(\textbf{W}_{i-1}\ast\left[\frac{\partial c_{n-i+1}}{\partial p};s_{i-1}\odot\frac{\partial\sigma(l_{i-2})}{\partial p} \right]\right)
\end{aligned}
$}
\end{equation}

\noindent This analysis on $\frac{\partial\sigma(l_{i-1})}{\partial p}$ applies recursively to $\frac{\partial\sigma(l_{i-2})}{\partial p}$; thus, $\frac{\partial^2G}{\partial z\partial p}\rightarrow 0$ if $\forall\,i.\,\frac{\partial c_i}{\partial p}\rightarrow 0$.

\begin{figure}[t]
   \centering
   \includegraphics[width=1\linewidth]{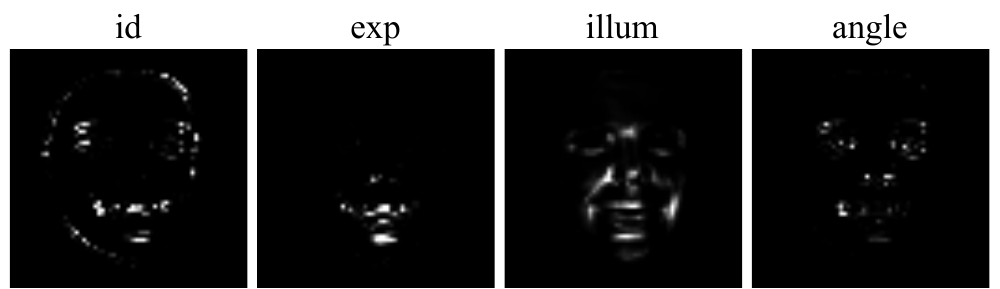}
   \vspace{-0.5cm} 
   \caption{Finite difference approximation of the partial derivative of the injected 3DMM render features~\wrt the 3DMM parameters $\frac{\partial c}{\partial p}$. We can see that the derivative maps are sparse, with the variation in $c$ depicted in small white regions, indicating that disentanglement is mostly successful.}
   \label{fig:3dmm_derivatives}
   \vspace{-0.3cm} 
\end{figure}

In practice, we empirically find that small variation in $p$ does lead to little total variation in $c$. Variation in $c$ tends to be highly localized to small affected regions dictated by $p$, with little variation otherwise (Fig.~\ref{fig:3dmm_derivatives}). This is likely the combination effect of localized variation in $\text{rep}$~\wrt $p$ and the inductive bias of locality of a convolutional encoder. 
We do not consider $\frac{\partial^2G}{\partial p\partial z}$ as disentanglement in this direction is automatically enforced by $\mathcal{L}_\text{consistency}$ when pairing each $p$ with a set of different $z$s.

\section{Experiments}

Following previous works~\cite{dfg,3fg}, we evaluate our methods on $256 \times 256$ FFHQ~\cite{sg1}. We compare our model against StyleGAN2 and two state-of-the-art 3DMM-based generative models, DiscoFaceGAN (DFG)~\cite{dfg} and 3D-FM GAN~\cite{3fg}. As the leading SOTA method 3D-FM GAN does not have public code or models, comparison is difficult. Where possible, we took results from their paper, but some quantitative metrics could only be computed for our model and for DiscoFaceGAN. 


{   

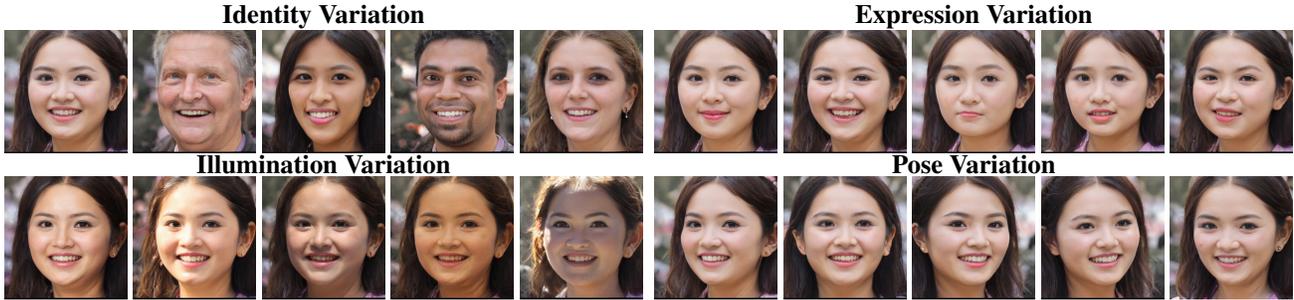
\begin{figure*}[t]
    \vspace{-0.5cm}
    \centering
    \setlength{\tabcolsep}{1pt}
    
    \newcommand{\pathOne}{figures/fig_qualitative_1/}
    \newcommand{\pathTwo}{figures/fig_qualitative_1/}
    \newcommand{\pathThree}{figures/fig_qualitative_1/}

    \newcommand{\rgbcolormapBrighter}{{0.2 1.2 0.2 1.2 0.2 1.2}}
    \newcommand{\rgbcolormap}{{0 1 0 1 0 1}}

    \newcommand{\imgw}{0.19\linewidth}
    \newcommand{\imgh}{0.19\linewidth}
    
    \newcommand{\imgRGB}[2][0px 0px 0px 0px]{\includegraphics[draft=\isdraft,width=\imgw,height=\imgh,trim=#1,clip]{#2}}

    
    \newcommand{\zoomRGB}[4]{
        \begin{tikzpicture}[
    		image/.style={inner sep=0pt, outer sep=0pt},
    		collabel/.style={above=9pt, anchor=north, inner ysep=0pt, align=center}, 
    		rowlabel/.style={left=9pt, rotate=90, anchor=north, inner ysep=0pt, scale=0.8, align=center},
    		subcaption/.style={inner xsep=0.75mm, inner ysep=0.75mm, below right},
    		arrow/.style={-{Latex[length=2.5mm,width=4mm]}, line width=2mm},
    		spy using outlines={rectangle, size=1.25cm, magnification=3, connect spies, ultra thick, every spy on node/.append style={thick}},
    		style1/.style={cyan!90!black,thick},
    		style2/.style={orange!90!black},
    		style3/.style={blue!90!black},
    		style4/.style={green!90!black},
    		style5/.style={white},
    		style6/.style={black},
        ]
        
        \node [image] (#1) {\imgRGB{#2}};
        \spy[style5] on ($(#1.center)-#3$) in node (crop-#1) [anchor=#4] at (#1.#4);
        
        \end{tikzpicture}
    }
    
    
    \begin{minipage}[t]{0.49\linewidth}
    \centering
    \textbf{Identity Variation}
    

    
    \imgRGB{\pathThree id/idx786_base.png}
    \imgRGB{\pathThree id/idx786_id123_1.0_expNone_1.0_gammaNone_1.0_xNone_yNone.png}   \imgRGB{\pathThree id/idx786_id1255_1.0_expNone_1.0_gammaNone_1.0_xNone_yNone.png}  \imgRGB{\pathThree id/idx786_id92_1.0_expNone_1.0_gammaNone_1.0_xNone_yNone.png}    \imgRGB{\pathThree id/idx786_id546_0.9_expNone_1.0_gammaNone_1.0_xNone_yNone.png}
    \end{minipage}
    %
    \begin{minipage}[t]{0.49\linewidth}
    \centering
    \textbf{Expression Variation}



    \imgRGB{\pathThree exp/idx786_idNone_1.0_exp1445_1.0_gammaNone_1.0_xNone_yNone.png}
    \imgRGB{\pathThree exp/idx786_idNone_1.0_exp142_1.74_gammaNone_1.0_xNone_yNone.png}
    \imgRGB{\pathThree exp/idx786_idNone_1.0_exp142_-0.69_gammaNone_1.0_xNone_yNone.png}
    \imgRGB{\pathThree exp/idx786_idNone_1.0_exp34310_1.69_gammaNone_1.0_xNone_yNone.png}
    \imgRGB{\pathThree exp/idx786_idNone_1.0_exp23333_1.43_gammaNone_1.0_xNone_yNone.png}
    
    \end{minipage}
    %
    
    \begin{minipage}{0.49\linewidth}
    \centering
    \textbf{Illumination Variation}
    

    
    \imgRGB{\pathThree illum/idx786_idNone_1.0_expNone_1.0_gamma23453_1.0_xNone_yNone.png}
    \imgRGB{\pathThree illum/idx786_idNone_1.0_expNone_1.0_gamma23444_1.0_xNone_yNone.png}
    \imgRGB{\pathThree illum/idx786_idNone_1.0_expNone_1.0_gamma4534_0.59_xNone_yNone.png}
    \imgRGB{\pathThree illum/idx786_idNone_1.0_expNone_1.0_gamma1813_0.46_xNone_yNone.png}
    \imgRGB{\pathThree illum/idx786_idNone_1.0_expNone_1.0_gamma41242_-0.16_xNone_yNone.png}

    \end{minipage}
    %
    \begin{minipage}{0.49\linewidth}
    \centering
    \textbf{Pose Variation}
    

    
    \imgRGB{\pathThree angle/idx786_idNone_1.0_expNone_1.0_gammaNone_1.0_xNone_y-0.4.png}
    \imgRGB{\pathThree angle/idx786_idNone_1.0_expNone_1.0_gammaNone_1.0_x0_y0.png}
    \imgRGB{\pathThree angle/idx786_idNone_1.0_expNone_1.0_gammaNone_1.0_xNone_y0.3.png}
    \imgRGB{\pathThree angle/idx786_idNone_1.0_expNone_1.0_gammaNone_1.0_x0.1_y0.25.png}
    \imgRGB{\pathThree angle/idx786_idNone_1.0_expNone_1.0_gammaNone_1.0_x0.2_y-0.25.png}

    \end{minipage}



    \vspace{-0.2cm}
    \caption{Generated face samples with control as output from our model. While some unwanted variation remains, identity, expression, illumination, and angle are controlled with high fidelity and no apparent visual artifacts.}
    \label{fig:qualitative_synthetic_one}

    \vspace{-0.4cm}
\end{figure*}

}

\vspace{-0.05cm}
\subsection{Qualitative Comparison}
\begin{figure}[t]
   \centering
   \includegraphics[width=1\linewidth]{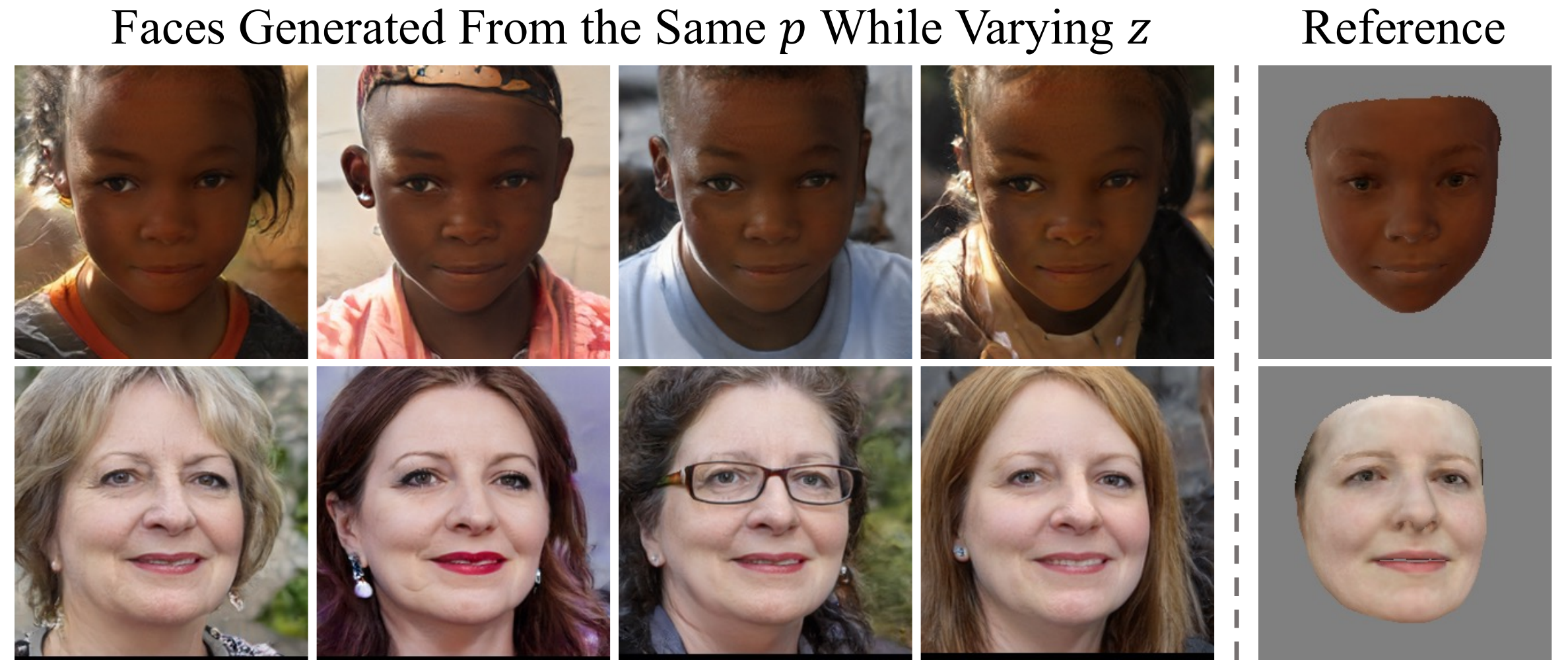}
    \vspace{-0.5cm}
   \caption{Resampling the noise vector $z$ with the same set of 3DMM coefficients $p$ shows high facial consistency, while other unsupervised factors such as hair, hat, eyeglasses, and background vary with $z$.}
   \label{fig:p_consistency}
\vspace{-15px}
\end{figure}

Our model achieves highly controllable generation while preserving StyleGAN's ability to generate highly photorealistic images (Fig.~\ref{fig:qualitative_synthetic_one}). We can see that our model can produce photorealistic faces with diverse races, genders, and ages. It also shows effective control over each of the 3DMM attributes. Particularly, we use the same person as our base image for all attribute edits; this verifies that our model can perform robust generation with high quality. Fig.~\ref{fig:p_consistency} compares the images generated by our model conditioned on the same $p$ but different $z$. The identity, expression, pose, and illumination are preserved while all other attributes can be modified. This means there is little overlap between attributes controlled by $p$ and $z$, and our model gains control over target attributes.



\subsection{Quantitative Comparison}

We evaluate the performance of our model in terms of quality and disentanglement. 
\vspace{-10px}
\paragraph{Fréchet inception distance (FID)}
For image quality, we compute the FID~\cite{fid} against the entire FFHQ dataset as a measure of the generation quality. Our model outperforms the two state-of-the-art baselines, yielding an FID much closer to the original StyleGAN trained on 256 $\times$ 256 FFHQ dataset (Tab.~\ref{tab:quantitative}).

\vspace{-10px}
\paragraph{Disentanglement Score (DS)} Introduced in DiscoFaceGAN, this quantifies the disentanglement efficacy of each of the four 3DMM-controlled attributes. For attribute vector $u_i \in \{z_{\text{id}}, z_{\text{exp}}, z_{\text{illum}}, z_{\text{angle}}\}$, we first randomly sample 1K sets of the other three attribute vectors, denoted by $u_{\{j\}} = \{u_j : j = 1, ..., 4, j \neq i\}$. Then, for each set of $u_{\{j\}}$, we randomly sample 10 $u_i$. In total, we have 10K 3DMM coefficients and hence generate 10K images. Then, we re-estimate $u_i$ and $u_{\{j\}}$ using the 3D reconstruction network ~\cite{recon}. For each attribute, we compute the L2 norm of the difference between each $u$ and the mean $u$ vector and get the mean L2 norm in each of the 1K sets. We then get $\sigma_{u_i}$ and $\sigma_{u_j}$'s by averaging the corresponding mean L2 norm over the 1K sets and normalize them by the L2 norm of the mean $u$ vector computed on the entire FFHQ dataset. Finally, we compute the disentanglement score:
\vspace{-5px}
\begin{align}
    DS(u_i) = \prod_{j, j \neq i} \frac{{\sigma}_{u_i}}{{\sigma}_{u_j}}
\end{align}
\vspace{-10px}

A high $DS$ indicates that when an attribute vector is modified, only the corresponding attribute is changed on the generated image while all other attributes remain unchanged. Our model outperforms DiscoFaceGAN by large margins in identity, expression, and pose (angle) control (Table \ref{tab:quantitative}).






\begin{table}[t]
    \caption{Our conditioning provides control and almost equivalent quality to unconditioned baseline StyleGAN2. The baseline 3DMM conditioning approaches do not produce comparable quality in terms of FID and DS. }
    \centering
    \resizebox{\linewidth}{!}{
    \begin{tabular}{ l|c|cccc }
        \toprule
        Method & FID$\downarrow$ & $DS_{\text{id}}\uparrow $ & $DS_{\text{exp}}\uparrow$ & $DS_{\text{illum}}\uparrow$ & $DS_{\text{angle}}\uparrow$ \\
        \midrule
        StyleGAN2 & 3.78 & - & - & - & - \\
        Ours & 4.51 & 1.02 & 3.22 & 48.7 & 1245 \\
        DiscoFaceGAN & 12.9 & 0.37 & 1.64 & 47.9 & 829 \\
        3D-FM GAN & 12.2 & - & - & - & -\\
        \bottomrule
    \end{tabular}
    }
    \label{tab:quantitative}
    \vspace{-10px}
\end{table}

\vspace{-0px}
\section{Conclusion and Future Work}
\vspace{-0px}
We present a simple conditional model derived from a mathematical framework for 3DMM conditioned face generation. Our model shows strong performance in both quality and controllability, reducing the need to choose between the two and making control `tax free'. Furthermore, our mathematical framework can be applied to future explorations in conditional generation, allowing future investigators to analyze other 3DMMs rigorously.
However, our model does not come without limitations. Unlike 3D-FM GAN~\cite{3fg}, our model is not specifically designed for image editing. Thus, faces suffer the same inversion accuracy~\vs editability tradeoff as StyleGAN~\cite{sg1,sg2,sg3}. Future work might consider applying the image editing techniques proposed by Liu~\etal~\cite{3fg} to our model for better face editing. 


{\small
\bibliographystyle{iccv_template/ieee_fullname}
\bibliography{bib/face_editing}
}








\crefname{section}{Sec.}{Secs.}
\Crefname{section}{Section}{Sections}
\Crefname{table}{Table}{Tables}
\crefname{table}{Tab.}{Tabs.}


\def\cvprPaperID{24} 
\def\httilde{\mbox{\tt\raisebox{-.5ex}{\symbol{126}}}}







\appendix

\clearpage
\section*{Appendix}

\section{Implementation Details}
We implement our model on top of the official StyleGAN2~\cite{sg2} and the PyTorch release of Deep3DRecon~\cite{recon}. $\textbf{FR}$ and $\textbf{RDR}$ are both part of Deep3DRecon~\cite{recon} and $G$ and $D$ are part of StyleGAN2~\cite{sg2}. We use the dataset tool provided in Deep3DRecon~\cite{recon} to realign FFHQ~\cite{sg1} so that image $x$ aligns with 3DMM representation $\text{rep}$.

\paragraph{StyleGAN2 backbone.}
We follow the latest findings in StyleGAN3~\cite{sg3} and omit several insignificant details to simplify StyleGAN2~\cite{sg2}. We remove mixing regularization and path length regularization. The depth of the mapping network is decreased to 2, as recommended by Karras~\etal. It is also noticed that decreasing the dimensionality of $z$ while maintaining the dimensions of $w$ is beneficial~\cite{sgxl}. Therefore, we reduce the dimensions of $z$ to 64. All details are otherwise unchanged, including the network architecture, equalized learning rate, minibatch standard deviation, weight (de)modulation, lazy regularization, bilinear resampling, and exponential moving average of the generator weights.

\paragraph{Face reconstruction and differentiable renderer.}
We use the pretrained checkpoint provided by Deng~\etal~\cite{recon} for $\textbf{FR}$. This updated checkpoint was trained on an augmented dataset that includes FFHQ~\cite{sg1} and shows slight performance improvement over the TensorFlow release of Deep3DRecon. We use the differentiable renderer $\textbf{RDR}$ that comes with the checkpoint for $\textbf{FR}$ from the same code repository. This renderer uses the Basel Face Model from 2009~\cite{bfm09} as the 3DMM parametric model for face modeling, and nvdiffrast~\cite{nvrast} for rasterization. We modify $\textbf{RDR}$ so it outputs $a$ and $n$ along with $r$. The renderer is otherwise unchanged.

\paragraph{Training procedure.}
Following the StyleGAN family~\cite{sg1,sg2,sg3}, we adopt the non-saturating loss~\cite{gan} and R1 gradient penalty~\cite{r1} as the loss function for GAN training. We additional append our $\mathcal{L}_\text{consistency}$, resulting in the following objectives:
\begin{align}
\mathcal{L}_D=-&\mathbb{E}_{p,z}\left[\log(1-D(G(\text{rep}(p),z)))\right] - \nonumber \\
&\mathbb{E}_{\substack{x}}\left[\log(D(x))\right]+\frac{\gamma}{2}\mathbb{E}_{\substack{x}}\left[\left \| \nabla D(x)\right \|^2_2\right]
\end{align}
\begin{align}
\mathcal{L}_G=-\mathbb{E}_{p,z}\left[\log(D(G(\text{rep}(p),z)))\right]+\lambda\mathcal{L}_\text{consistency}
\end{align}

\noindent We closely follow the training configurations of the baseline model in Karras et al.~\cite{sg2ada} and set $\gamma=1$. The batch size is set to 64 and the group size of minibatch standard deviation is set to 8. We empirically set $\lambda=20$ and the length of progressive blending to $k=2\times10^6$. The learning rate of both $G$ and $D$ is set to $2.5\times10^{-3}$. We train our model until $D$ sees 25M real images~\cite{sg1,sg2,sg3}.

Instead of approximating the distribution $P(p)$ using a VAE~\cite{dfg}, we simply use its empirical distribution when sampling $p\sim P(p)$ and find this to be sufficient given our 3DMM representation.

\begin{figure}[t]
   \centering
   \includegraphics[width=\linewidth]{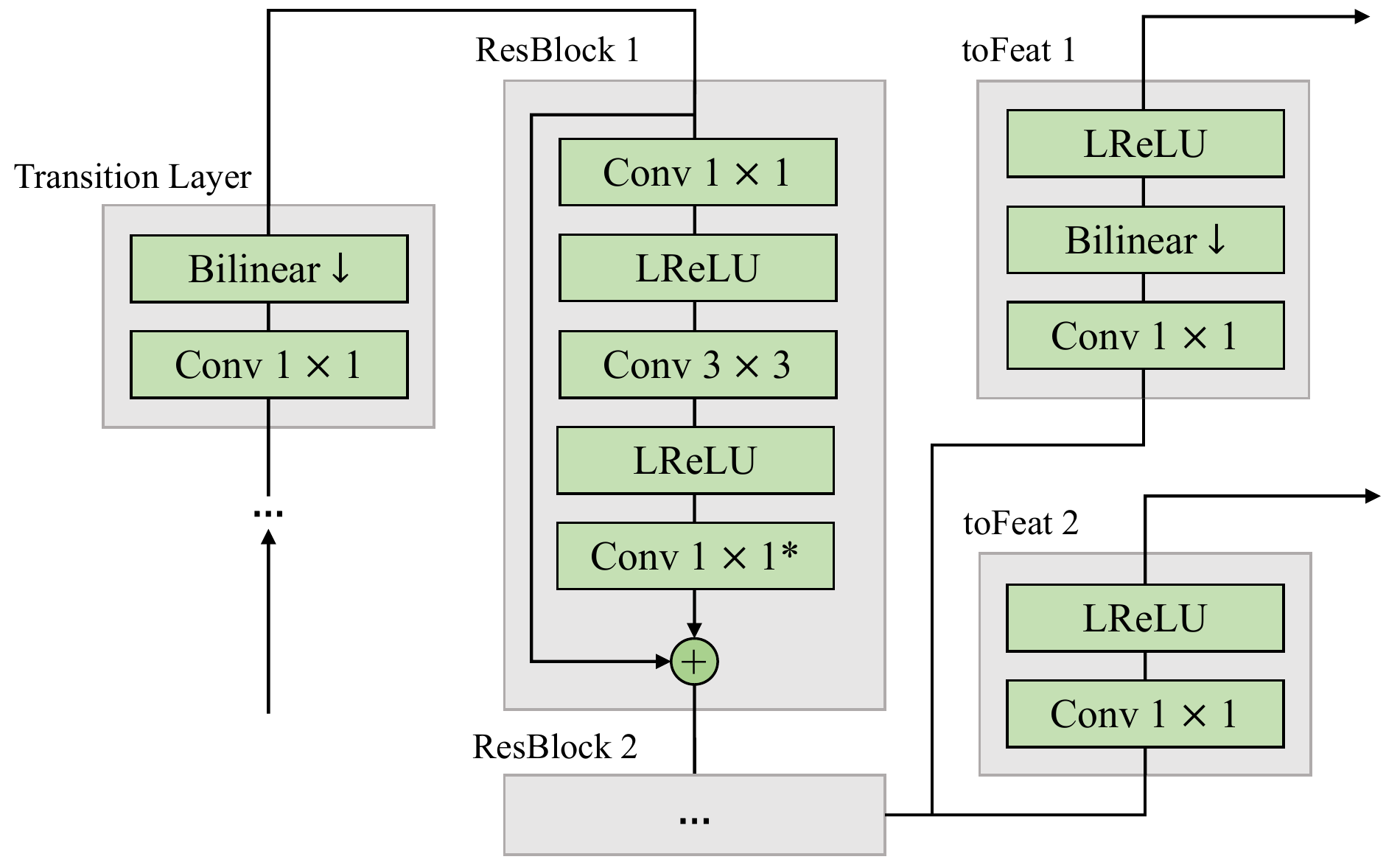}
   \caption{The detailed breakdown of a general stage of $E$.}
   \label{fig:encoderdetail}\vspace{-0.35cm}
\end{figure}

\section{Encoder Architecture}
We follow the architecture design of StyleGAN2 and split $E$ into different resolution stages. For each resolution stage $e_i$ of $E$, %
we produce two sets of feature maps $c_{2i}$ and $c_{2i+1}$ to condition the two synthesis layers of the corresponding resolution stage of the synthesis network:

\begin{equation}
\begin{aligned}[t]
&e_i=
\begin{cases}
E_0(\text{rep}(p))\,\,\,\,\,i=0 \\
E_i(e_{i-1})\,\,\,\,\,\,\,\,\,\,i\neq 0 \\
\end{cases} \\
&c_{2i}=\text{toFeat}_{2i}(e_i) \\
&c_{2i+1}=\text{toFeat}_{2i+1}(e_i)
\end{aligned}
\end{equation}

\noindent We implement $E_i$ as a sequence of a transition layer and two residual blocks (Fig.~\ref{fig:encoderdetail}). `$\text{toFeat}$' is implemented by a $1\times1$ convolution~\cite{cnn} with optional downsampling~\cite{sg1} and leaky ReLU activation~\cite{leakyrelu}. Following recent advances in network architecture~\cite{convnext, metaformer}, our ResNet~\cite{resnet} design of $E$ differs from the architecture of $D$~\cite{sg2} in several ways.

\paragraph{General stage.}
We notice that the two architectural changes in~\cite{convnext} that lead to most performance boost are separate downsampling layers and fewer activations. Thus, we move the skip branch of the transition residual block up to the stem as a transition layer, and remove all activations in the residual block unless they are between two consecutive convolutional layers. We use leaky ReLU activation with $\alpha=0.2$, and bilinear downsampling instead of strided convolution~\cite{sg1,sg2}. We use the 1-3-1 bottleneck residual block as it is more efficient than the 3-3 block~\cite{resnet}. The final convolutional layer (marked by *) in the residual block is initialized to 0~\cite{fixup}, and this eliminates the need for normalization or residual rescaling~\cite{sg2}. We apply equalized learning rate to all convolutional layers.

\paragraph{Specialization.} We remove bilinear downsampling from the transition layer of the highest resolution stage; it is otherwise identical to a general stage. Since the $4\times4$ stage of the synthesis network contains only one synthesis layer, we place one $\text{toFeat}$ layer without leaky ReLU in the $4\times4$ stage of $E$ accordingly.

\begin{figure}[t]
   \centering
   \includegraphics[width=\linewidth]{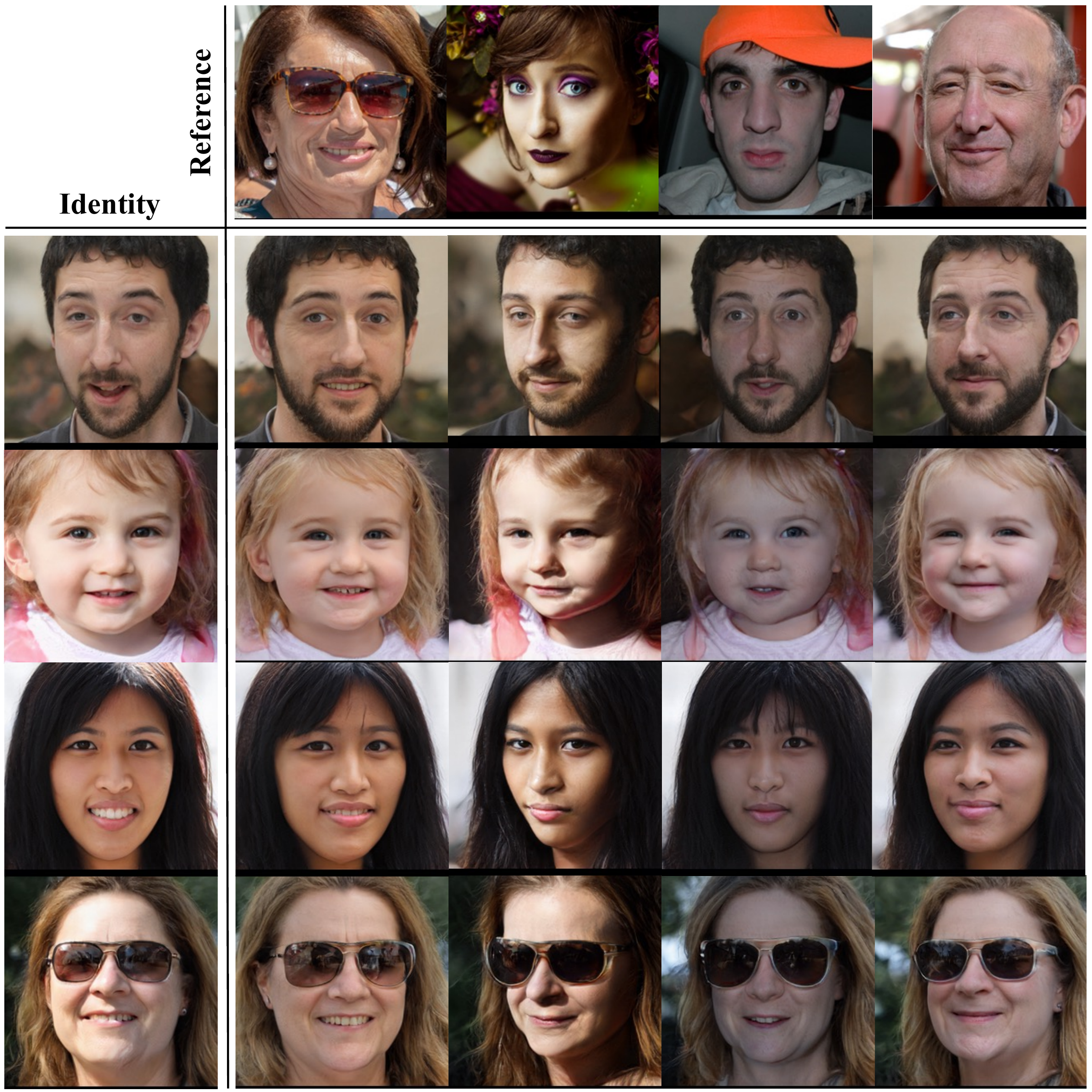}
   \caption{Reference-based generation results. We extract the expression, illumination, and pose coefficients from the reference images (first row) and apply them to randomly generated images (first column).}
   \label{fig:refbased}\vspace{-0.35cm}
\end{figure}

\begin{figure}[t]
   \centering
   \includegraphics[width=\linewidth]{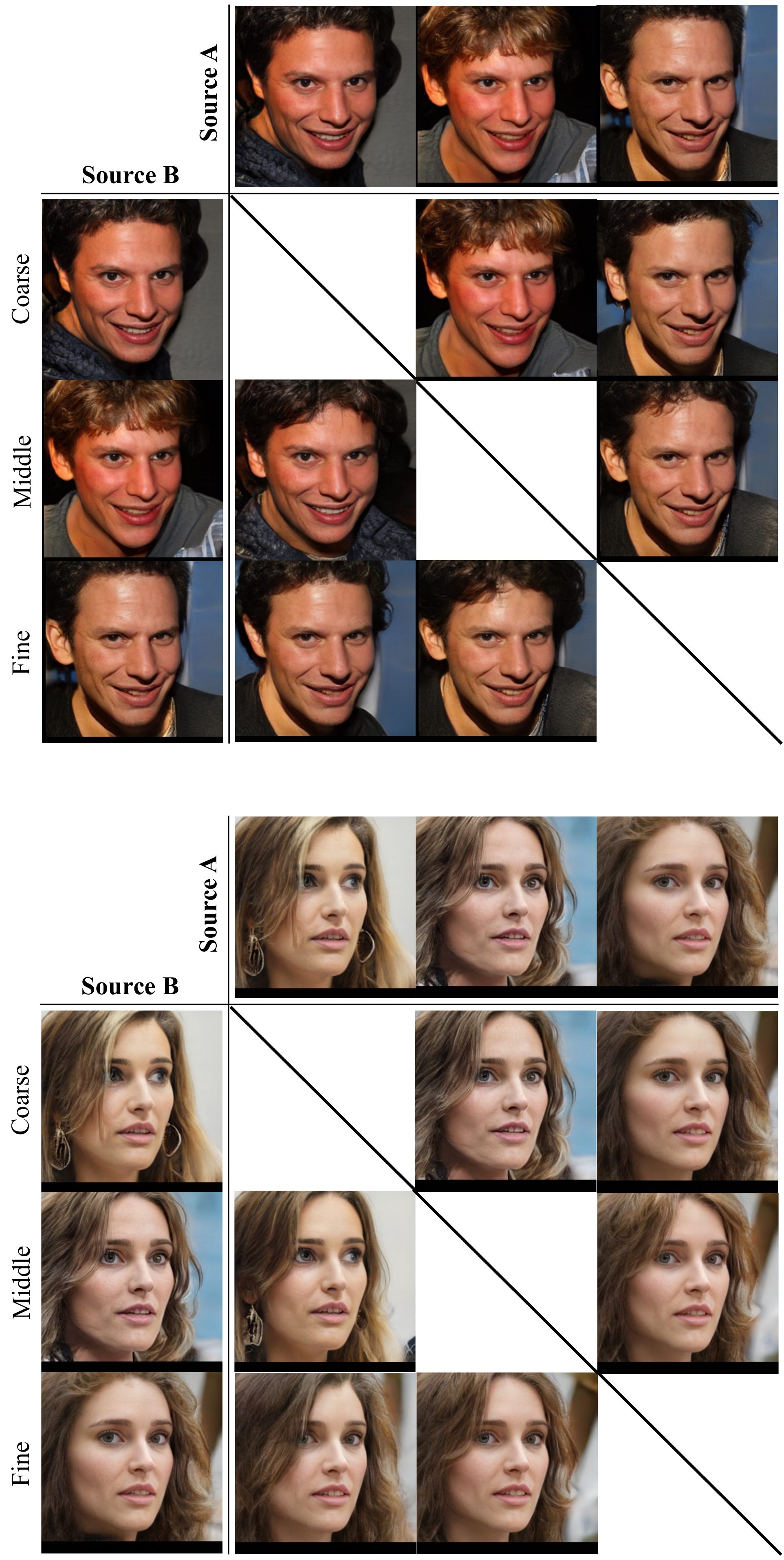}
   \caption{Style mixing results at different scales. Using the same three images for Source A and Source B, we replace the style vectors of images from Source A by the style vectors of images from Source B at coarse resolutions (4$\times$4 - 8$\times$8), middle resolutions (16$\times$16 - 32$\times$32), and fine resolutions (64$\times$64 - 256$\times$256).}
   \label{fig:wMix}\vspace{-0.35cm}
\end{figure}

\begin{figure}[t]
   \centering
   \includegraphics[width=\linewidth]{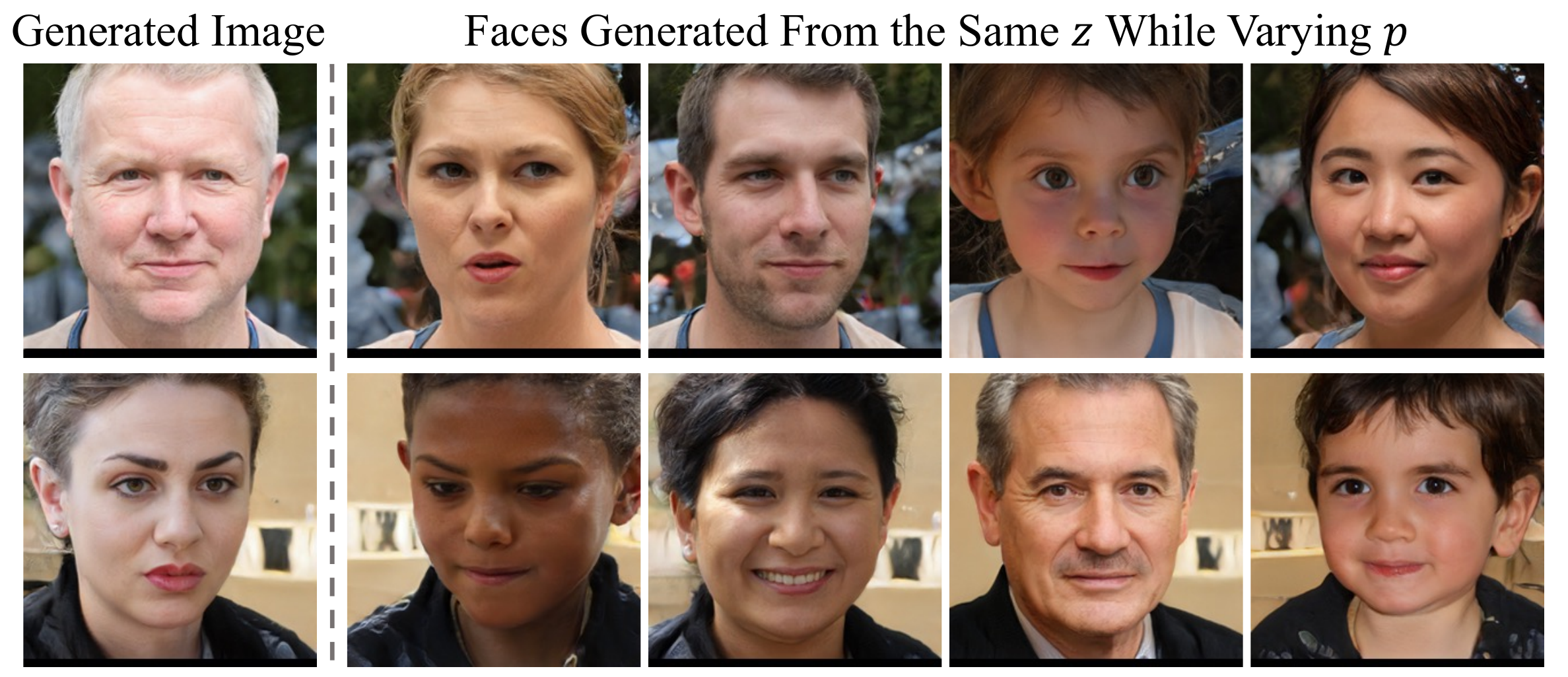}
   \caption{Resampling the 3DMM coefficient vector $p$ with the same noise vector $z$ shows high consistency in the background and clothes while the face completely changes.}
   \label{fig:zConsistency}\vspace{-0.35cm}
\end{figure}

\section{More Results}
We show additional results in controlled generation that display the robustness of our model and explain what control exists in the non-conditioned $z$ space.

\paragraph{Reference-based generation}
In Fig.~\ref{fig:refbased}, we task our model with reference-based generation where we keep the identity of a generated image and swap its expression, illumination, and pose with those of a real image. We can see that the respective attributes from their source are all well preserved, and the image quality does not degrade. This again demonstrates the disentangled face generation from our model.

\paragraph{Feature granularity} 
To inspect the impact of feature variability across the layers of the decoder, we inspect the impact of swapping features across images with the same $p$. In Fig.~\ref{fig:wMix}, we randomly pick a 3DMM coefficient vector $p$ and randomly sample $z$'s to generate three images (the same images for Source A and Source B). Following StyleGAN~\cite{sg1}, we replace some of the style vectors $w^+$ of images from Source A by the corresponding style vectors of images from Source B at coarse, middle, and fine scales. As $p$ is the same, the overall face region will not change significantly.

At coarse scale, there is no visible change to the images from Source A. This is expected as the high-level attributes of the image are supposed to be determined by the $p$ vector. At middle scale, the images from Source A remain mostly unchanged except finer facial features such as the hair now resemble those in the image from Source B. At fine scale, the images from Source A undergo more significant changes where the color scheme that affects the background, clothes, hair color, and skin color now resembles those in the image from Source B. This experiment indicates that each subset of the style vectors $w^+$ controls a different set of features in the generated image. We also notice that attributes controlled by $p$ remain unchanged at any scale, which means our model's $p$ space and $z$ space are well separated.

\paragraph{3DMM vector resampling with fixed noise}
As opposed to the experiment conducted in the main paper where we vary the noise $z$ with fixed 3DMM vector $p$, we now vary $p$ with fixed $z$ as in Fig. \ref{fig:zConsistency}. We can see that despite the drastic change in the facial attributes from different $p$'s, the background and clothes remain largely consistent with the same $z$. This is another proof that the $z$ vector has a good control of the attributes not controlled by $p$.  

\begin{figure}[t]
   \centering
   \includegraphics[width=\linewidth]{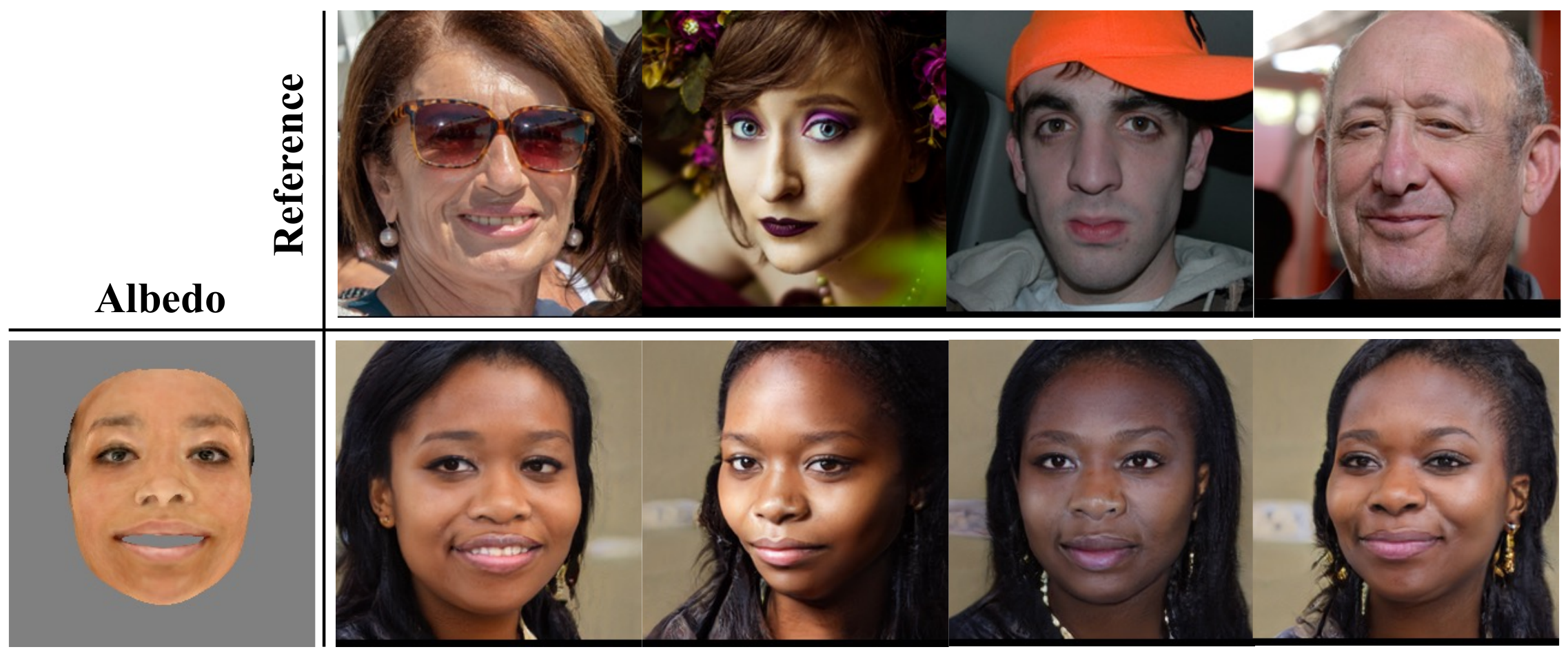}
   \caption{Reference-based generation results that show unexpected skin tone change. We see that the albedo predicted by $\textbf{FR}$ does not faithfully capture the darker skin tone.}
   \label{fig:limit}\vspace{-0.35cm}
\end{figure}

\paragraph{Limitations.}
Due to the use of a pretrained $\textbf{FR}$ and $\textbf{RDR}$, our model inevitably inherits the limitations of these models. We find that Deep3DRecon~\cite{recon} performs particularly poor on darker skin tone, in that it tends to predict the skin tone as the result of dim illumination. This leads to unexpected skin tone change when editing the illumination (Figure.\ref{fig:limit}). Moreover, our model does not provide explicit control over attributes not represented in $\mathcal{P}$ such as hair and eyeglasses. We believe these restrictions can be resolved in the future by an improved 3DMM.

\end{document}